\documentclass[runningheads]{llncs}
\usepackage{amsmath,graphicx, amssymb, amsfonts, bbm}
\usepackage{bm}
\usepackage{color}
\usepackage{algorithmic}
\usepackage{graphicx}
\usepackage{textcomp}

\usepackage{soul}
\usepackage{url}
\usepackage[utf8]{inputenc}
\usepackage[small]{caption}
\usepackage{booktabs}
\usepackage{algorithm}
\usepackage{algorithmic}
\usepackage{calc}

\usepackage{multirow}
\usepackage{rotating}
\usepackage{balance}

\makeatletter
\def\Hline{%
\noalign{\ifnum0=`}\fi\hrule \@height 1pt \futurelet
\reserved@a\@xhline}
\makeatother

\begin{document}
\title{Bridging the Gap between AI and Healthcare Sides: Towards Developing Clinically Relevant AI-Powered Diagnosis Systems}
\titlerunning{Bridging the Gap between AI and Healthcare Sides}
%
%
\author{Changhee Han\inst{1,2,3}\orcidID{0000-0002-4429-3859} \and
\\Leonardo Rundo\inst{4,5}\orcidID{0000-0003-3341-5483} \and
Kohei Murao\inst{3} \and
\\Takafumi Nemoto\inst{6} \and
Hideki Nakayama\inst{1}}
\authorrunning{C. Han et al.}
%
\institute{Graduate School of Information Science and Technology,\\The University of Tokyo, Tokyo, Japan \and
LPIXEL Inc., Tokyo, Japan\\
\email{han@lpixel.net} \and
Research Center for Medical Big Data,\\National Institute of Informatics, Tokyo, Japan \and
Department of Radiology, University of Cambridge, Cambridge, UK \and
Cancer Research UK Cambridge Centre, University of Cambridge, Cambridge, UK \and
Department of Radiology, Keio University School of Medicine, Tokyo, Japan}
\maketitle              
%
\begin{abstract}
Despite the success of Convolutional Neural Network-based Computer-Aided Diagnosis research, its clinical applications remain challenging. Accordingly, developing medical Artificial Intelligence (AI) fitting into a clinical environment requires identifying/bridging the gap between AI and Healthcare sides. Since the biggest problem in Medical Imaging lies in data paucity, confirming the clinical relevance for diagnosis of research-proven image augmentation techniques is essential. Therefore, we hold a clinically valuable AI-envisioning workshop among Japanese Medical Imaging experts, physicians, and generalists in Healthcare/Informatics. Then, a questionnaire survey for physicians evaluates our pathology-aware Generative Adversarial Network (GAN)-based image augmentation projects in terms of Data Augmentation and physician training. The workshop reveals the intrinsic gap between AI/Healthcare sides and solutions on \textit{Why} (i.e., clinical significance/interpretation) and \textit{How} (i.e., data acquisition, commercial deployment, and safety/feeling safe). This analysis confirms our pathology-aware GANs' clinical relevance as a clinical decision support system and non-expert physician training tool. Our findings would play a key role in connecting inter-disciplinary research and clinical applications, not limited to the Japanese medical context and pathology-aware GANs.

\end{abstract}
\section{Introduction}
\label{sec:1}
Convolutional Neural Networks (CNNs) have enabled accurate/reliable Computer-Aided Diagnosis (CAD), occasionally outperforming expert physicians~\cite{hwang2018development,wu2019TMI,mckinney2020}. However, such research results cannot be easily transferred to a clinical environment. Artificial Intelligence (AI) and Healthcare sides have a huge gap around technology, funding, and people~\cite{allen2019}. In Japan, the biggest challenge lies in medical data sharing because each hospital has different ethical codes and tends to enclose collected data without annotating them for AI research. This differs from the US, where National Cancer Institute provides annotated medical images~\cite{Clark13}.

Therefore, a Research Center for Medical Big Data was launched in November 2017: collaborating with 6 Japanese medical societies and 6 institutes of informatics, we collected large-scale annotated medical images for CAD research. Using over 60 million available images, we achieved prominent research results, presented at major Computer Vision~\cite{Tokunaga19} and Medical Imaging conferences~\cite{Kanayama19}. Moreover, we published 6 papers~\cite{Han1,Han2,han2019combining,han2019learning,han2019synthesizing,han2019learning2} on Generative Adversarial Network (GAN)-based medical image augmentation~\cite{Goodfellow}. Since the GANs can generate realistic samples with desired pathological features \textit{via} many-to-many mappings, they could mitigate the medical data paucity \textit{via} Data Augmentation (DA) and physician training.

\begin{figure}[t!]
  \centering
  \centerline{\includegraphics[width=0.72\linewidth]{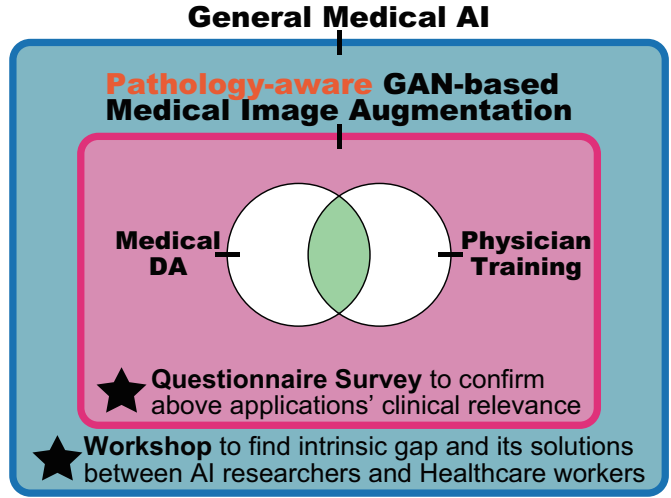}}
\caption{Overview of our discussions towards developing clinically relevant AI-powered diagnosis systems: (\textit{i}) A workshop for 7 Japanese people with various AI and/or Healthcare background to develop medical AI fitting into a clinical environment in five years; (\textit{ii}) A questionnaire survey for 9 Japanese physicians to confirm our pathology-aware GAN-based realistic/diverse image augmentation's clinical relevance---the medical DA requires high diversity whereas the physician training requires high realism.}
\label{fig1}
\vspace{-0.4cm}
\end{figure}

Aiming to further identify/bridge the gap between AI and Healthcare sides in Japan towards developing medical AI fitting into a clinical environment in five years, we hold a workshop for $7$ Japanese people with various AI and/or Healthcare background. Moreover, to confirm the clinical relevance for diagnosis of the pathology-aware GAN methods, we conduct a questionnaire survey for $9$ Japanese physicians who interpret Computed Tomography (CT) and Magnetic Resonance (MR) images in daily practice. Fig.~\ref{fig1} outlines our investigation\vspace{0.1cm}.

\noindent \textbf{Contributions.} Our main contributions are as follows:
\vspace{-0.1cm}
\begin{itemize}
    \item \textbf{AI and Healthcare Workshop}: We firstly hold a clinically valuable AI-envisioning workshop among Medical Imaging experts, physicians, and Healthcare/Informatics generalists to bridge the gap between AI/Healthcare sides.
    \item \textbf{Questionnaire Survey for Physicians}: We firstly present both qualitative/quantitative questionnaire evaluation results for many physicians about research-proven medical AI.
    \item \textbf{Information Conversion}: Clinical relevance discussions imply that our pathology-aware GAN-based interpolation and extrapolation could overcome medical data paucity \textit{via} DA and physician training.
\end{itemize}

\vspace{-0.4cm}

\vspace{-0.2cm}
\section{Pathology-aware GAN-based Image Augmentation}
\label{sec:2}
\vspace{-0.15cm}

In terms of interpolation, GAN-based medical image augmentation is reliable because acquisition modalites (e.g., X-ray, CT, MR) can display the human body's strong anatomical consistency at fixed position while clearly reflecting inter-subject variability~\cite{hsieh2009computed,brown2014magnetic}. This is different from natural images, where various objects can appear at any position; accordingly, to tackle large inter-subject/pathology/modality variability, we proposed to use noise-to-image GANs (e.g., random noise samples to diverse pathological images) for (\textit{i}) medical DA and (\textit{ii}) physician training~\cite{Han1}. While the noise-to-image GAN training is much more difficult than training image-to-image GANs~\cite{tmenova2019cyclegan} (e.g., a benign image to a malignant one), it can increase image diversity for further performance boost.

Regarding the DA, the GAN-generated images can improve CAD based on supervised learning~\cite{frid2018gan,Madani,konidaris2018generative}. For the physician training, the GANs can display novel desired pathological images and help train medical trainees despite infrastructural/legal constraints~\cite{finlayson2018towards}. However, we have to devise effective loss functions and training schemes for such applications. Diversity matters more for the DA to sufficiently fill the real image distribution, whereas realism matters more for the physician training not to confuse the trainees.


So, how can we perform GAN-based DA/physician training with only 
limited annotated images? Always in collaboration with physicians, for improving 2D classification, we combined the noise-to-image and image-to-image GANs~\cite{Han2,han2019combining}.
Nevertheless, further DA applications require pathology localization for detection and advanced physician training needs the generation of images with abnormalities, respectively. To meet both clinical demands, we proposed 2D/3D bounding box-based GANs conditioned on pathology position/size/appearance. Indeed, the bounding box-based detection requires much less physicians' annotation effort than segmentation \cite{rundo2018NC}.


In terms of extrapolation, the pathology-aware GANs are promising because common and/or desired medical priors can play a key role in the conditioning---theoretically, infinite conditioning instances, external to the training data, exist and enforcing such constraints have an extrapolation effect \textit{via} model reduction~\cite{stinis2019enforcing}. For improving 2D detection, we proposed Conditional Progressive Growing of GANs that incorporates rough bounding box conditions incrementally into a noise-to-image GAN (i.e., Progressive Growing of GANs~\cite{Karras}) to place realistic/diverse brain metastases at desired positions/sizes on $256 \times 256$ MR images~\cite{han2019learning}. Since the human body is 3D, for improving 3D detection, we proposed 3D Multi-Conditional GAN that translates noise boxes into realistic/diverse $32 \times 32 \times 32$ lung nodules \cite{alShabi2019lung} placed at desired position/size/attenuation on CT scans~\cite{han2019synthesizing}. Interestingly, inputting the noise box with the surrounding tissues has the effect of combining the noise-to-image and image-to-image GANs.

We succeeded to (\textit{i}) generate images even realistic for physicians and (\textit{ii}) improve detection using synthetic training images, respectively; they require different loss functions and training schemes. However, to exploit our pathology-aware GANs as a (\textit{i}) non-expert physician training tool and (\textit{ii}) clinical decision support system, we need to confirm their clinical relevance for diagnosis---such information conversion~\cite{hondamulti} techniques to overcome the data paucity, not limited to our pathology-aware GANs, would become a clinical breakthrough.

\vspace{-0.25cm}
\section{Methods}
\label{sec:3}
\vspace{-0.15cm}
\subsection{AI and Healthcare Workshop}
\vspace{-0.05cm}
\label{sec:3.1}

\begin{itemize}
\item \textbf{Subjects:} $2$ Medical Imaging experts (i.e., a Medical Imaging researcher and a medical AI startup entrepreneur), $2$ physicians (i.e., a radiologist and a psychiatrist), and $3$ Healthcare/Informatics generalists (i.e., a nurse and researcher in medical information standardization, a general practitioner and researcher in medical communication, and a medical technology manufacturer's owner and researcher in health disparities)\\
\begin{table*}[t!]
\caption[Workshop program to (\textit{i}) know the overview of Medical Image Analysis and (\textit{ii}) find the intrinsic gap and solutions between AI researchers/Healthcare workers.]{Workshop program to (\textit{i}) know the overview of Medical Image Analysis and (\textit{ii}) find the intrinsic gap and its solutions between AI researchers/Healthcare workers. \textbf{*} indicates activities given by a facilitator (i.e., the first author), such as lectures.}
\label{tab8_1}
\centering
\scalebox{0.73}
{
\begin{tabular}{ll}
\Hline\noalign{\smallskip}
{\bfseries Time} (mins) & \bfseries Activity \\\noalign{\smallskip}\hline\noalign{\smallskip}
& \bfseries Introduction \\
10 & 1. Explanation of the workshop's purpose and flow\textbf{*} \\
10 & 2. Self-introduction and explanation of motivation for participation \\
5 & 3. Grouping into two groups based on background\textbf{*}\\
\noalign{\smallskip}\hline\noalign{\smallskip}
& \bfseries Learning: Knowing Medical Image Analysis \\
15 & 1. TED speech video watching: \textit{Artificial Intelligence Can Change} \textit{the future of Medical Diagnosis}\textbf{*}\\
35 & 2. Lecture: Overview of Medical Image Analysis including state-of-the-art research, well-known\\ &  challenges/solutions, and the summary of our pathology-aware GAN projects\textbf{*}\\ &  (its video in Japanese: \url{https://youtu.be/rTQLknPvnqs})\\
10 & 3. Sharing expectations, wishes, and worries about Medical Image Analysis \\ &  (its video in Japanese: \url{https://youtu.be/ILPEGga-hkY})\\
10 & Intermission\\
\noalign{\smallskip}\hline\noalign{\smallskip}
& \bfseries Thinking: Finding How to Develop Robust Medical AI\\
25 & 1. Identifying the intrinsic gap between AI/Healthcare sides after sharing their common and different \\ &  thinking/working styles\\
60 & 2. Finding how to develop gap-bridging medical AI fitting into a clinical environment in five years\\
10 & Intermission\\
\noalign{\smallskip}\hline\noalign{\smallskip}
& \bfseries Summary \\
25 & 1. Presentation\\
10 & 2. Sharing workshop impressions and ideas to apply obtained knowledge \\ &  (its video in Japanese: \url{https://youtu.be/F31tPR3m8hs})\\
5 & 3. Answering a questionnaire about satisfaction and further comments\\
5 & 4. Closing remarks\textbf{*}\\
\noalign{\smallskip}\Hline\noalign{\smallskip}
\end{tabular}
}
\vspace{-0.6cm}
\end{table*}
\vspace{-0.17cm}
\item \textbf{Experiments:} As its program shows (Table~\ref{tab8_1}), during the workshop, we conduct 2 activities: (\textit{Learning}) Know the overview of Medical Image Analysis, including state-of-the-art research, well-known challenges/solutions, and the summary of our pathology-aware GAN projects; (\textit{Thinking}) Find the intrinsic gap and its solutions between AI researchers and Healthcare workers after sharing their common and different thinking/working styles. This workshop was held on March 17th, 2019 at Nakayama Future Factory, Open Studio, The University of Tokyo, Tokyo, Japan.
\end{itemize}
\vspace{-0.6cm}
\subsection{Questionnaire Survey for Physicians}
\label{sec:3.2}
\vspace{-0.05cm}
\begin{itemize}
\item \textbf{Subjects:} $3$ physicians (i.e., a radiologist, a psychiatrist, and a physiatrist) committed to (at least one of) our pathology-aware GAN projects and $6$ project non-related radiologists without much AI background. This paper's authors are surely not included.\\
\vspace{-0.25cm}
\item \textbf{Experiments:} Physicians are asked to answer the following questionnaire within 2 weeks from December 6th, 2019 after reading 10 summary slides written in Japanese about general Medical Image Analysis and our pathology-aware GAN projects along with example synthesized images. We conduct both qualitative (i.e., free comments) and quantitative (i.e., five-point Likert scale~\cite{allen2007likert}) evaluation: Likert scale 1 $=$ very negative, 2 $=$ negative, 3 $=$ neutral, 4 $=$ positive, 5 $=$ very positive.\\
\item \textbf{Question 1:} Are you keen to exploit medical AI in general when it achieves accurate and reliable performance in the near future? (five-point Likert scale) Please tell us your expectations, wishes, and worries (free comments).
\item \textbf{Question 2:} What do you think about using GAN-generated images for DA? (five-point Likert scale)
\item \textbf{Question 3:} What do you think about using GAN-generated images for physician training? (five-point Likert scale)
\item \textbf{Question 4:} Any comments/suggestions about our projects towards developing clinically relevant AI-powered systems based on your experience?
\end{itemize}

\section{Results}
\label{sec:4}
\subsection{Workshop Results}
\label{sec:4.1}
\noindent We show the clinically-relevant findings from this Japanese workshop.\\ \ \\
\textbf{Gap between AI and Healthcare Sides}\\ \ \\
\noindent \textbf{Gap 1:} AI, including Deep Learning, provides unclear decision criteria, does it make physicians reluctant to use it for diagnosis in a clinical environment?

\begin{itemize}
\item \textbf{Healthcare side}: We rather expect applications other than diagnosis. If we use AI for diagnosis, instead of replacing physicians, we suppose a \textit{reliable second opinion}, such as alert to avoid misdiagnosis, based on various clinical data not limited to images---every single diagnostician is anxious about their diagnosis. AI only provides minimum explanation, such as a heatmap showing attention, which makes persuading not only the physicians but also patients difficult. Thus, the physicians' intervention is essential for intuitive explanation. Methodological safety and feeling safe are different. In this sense, pursuing explainable AI generally decreases AI's diagnostic accuracy~\cite{adadi2018peeking}, so physicians should still serve as mediators by engaging in high-level conversation or interaction with patients. Moreover, according to the medical law in most countries including Japan, only doctors can make the final decision. The first autonomous AI-based diagnosis without a physician was cleared by the Food and Drug Administration in 2018~\cite{abramoff2018pivotal}, but such a case is exceptional.\\
\vspace{-0.2cm}
\item \textbf{AI side}: Compared with other systems or physicians, Deep Learning's explanation is not particularly poor, so we require too severe standards for AI; the word \textit{AI} is excessively promoting anxiety and perfection. If we could thoroughly verify the reliability of its diagnosis against physicians by exploring uncertainty measures~\cite{nair2020exploring}, such intuitive explanation would be optional.
\end{itemize}

\noindent \textbf{Gap 2:} Are there any benefits to actually introducing medical AI?
\vspace{-0.05cm}
\begin{itemize}
\item \textbf{Healthcare side}: After all, even if AI can achieve high accuracy and convenient operation, hospitals would not introduce it without any commercial benefits. Moreover, small clinics, where physicians are desperately needed, often do not have CT or MR scanners~\cite{jankharia2008commentary}.\\
\vspace{-0.2cm}
\item \textbf{AI side}: The commercial deployment of medical AI is strongly tied to diagnostic accuracy; so, if it can achieve significantly outstanding accuracy at various tasks in the near future, patients would not visit hospitals/clinics without AI. Accordingly, introducing medical AI would become profitable in five years.
\end{itemize}

\noindent \textbf{Gap 3:} Is medical AI's diagnostic accuracy reliable?
\vspace{-0.05cm}
\begin{itemize}
\item \textbf{Healthcare side}: To evaluate AI's diagnostic performance, we should consider many metrics, such as sensitivity and specificity. Moreover, its generalization ability for medical data highly relies on inter-scanner/inter-individual variability~\cite{oConnor2017healthy}. How can we evaluate whether it is suitable as a clinically applicable system?\\
\vspace{-0.2cm}
\item \textbf{AI side}: Generally, alleviating the risk of overlooking the diagnosis is the most important, so sensitivity matters more than specificity unless their balance is highly disturbed. Recently, such research on medical AI that is robust to different datasets is active~\cite{Rundo2}.
\end{itemize}
\newpage
\noindent \textbf{How to Develop Medical AI Fitting into a Clinical Environment in Five Years}\\

\noindent \textbf{Why:} Clinical significance/interpretation
\vspace{-0.1cm}
\begin{itemize}
\item \textbf{Challenges}: We need to clarify which clinical situations actually require AI introduction. Moreover, AI's early diagnosis might not be always beneficial for patients.\\
\vspace{-0.2cm}
\item \textbf{Solutions}: Due to nearly endless disease types and frequent misdiagnosis coming from physicians' fatigue, we should use it as alert to avoid misdiagnosis~\cite{vandenberghe2017relevance} (e.g., reliable second opinion), instead of replacing physicians. It should help prevent oversight in diagnostic tests not only with CT and MR, but also with blood data, chest X-ray, and mammography before taking CT and MR~\cite{li2019medical}. It could be also applied to segmentation for radiation therapy~\cite{agn2016generative}, neurosurgery navigation~\cite{abi2018machine}, and pressure ulcers' echo evaluation. Along with improving the diagnosis, it would also make the physicians' workflow easier, such as by denoising~\cite{yang2018low}. Patients should decide whether they accept AI-based diagnosis under informed consent.

\end{itemize}

\noindent \textbf{How:} Data acquisition
\vspace{-0.1cm}
\begin{itemize}
\item \textbf{Challenges}: Ethical screening in Japan is exceptionally strict, so acquiring and sharing large-scale medical data/annotation are challenging---it also applies to Europe due to General Data Protection Regulation~\cite{rumbold2017effect}. Considering the speed of technological advances in AI, adopting it for medical devices is difficult in Japan, unlike in medical AI-ready countries, such as the US, where the ethical screening is relatively loose in return for the responsibility of monitoring system stability. Moreover, whenever diagnostic criteria changes, we need further reviews and software modifications. For example, the Tumor-lymph Node-Metastasis (TNM) classification~\cite{sobin2011tnm} criteria changed for oropharyngeal cancer in 2018 and for lung cancer in 2017, respectively. Diagnostic equipment/target changes also require large-scale data/annotation acquisition again.\\
\vspace{-0.2cm}
\item \textbf{Solutions}: 
For Japan to keep pace, the ethical screening should be adequate to the other leading countries. Currently, overseas research and clinical trials are proceeding much faster, so it seems better to collaborate with overseas companies than to do it in Japan alone. Moreover, complete medical checkup, which is extremely costly, is unique in East Asia, thus Japan could be superior in individuals' multiple medical data---Japan is the only country, where most workers aged 40 or over are required to have medical checkups once a year regardless of their health conditions by the Industrial Safety and Health Act~\cite{nawata2016evaluation}. To handle changes in diagnostic criteria/equipment and overcome dataset/task dependency, it is necessary to establish a common database creation workflow~\cite{mansour2019visual} by regularly entering electronic medical records into the database. For reducing data acquisition/annotation cost, AI techniques, such as GAN-based DA~\cite{han2019synthesizing} and domain adaptation~\cite{ghafoorian2017transfer}, would be effective.
\end{itemize}

\noindent \textbf{How:} Commercial deployment
\vspace{-0.1cm}
\begin{itemize}
\item \textbf{Challenges}: Hospitals currently do not have commercial benefits to actually introduce medical AI. \\
\vspace{-0.2cm}
\item \textbf{Solutions}: For example, it would be possible to build AI-powered hospitals~\cite{chen2019feasibility} operated with less staff. Medical manufacturers could also standardize data format~\cite{laplante2016hearing}, such as for X-ray, and provide some AI services. Many IT giants like Google are now working on medical AI to collect massive biomedical datasets~\cite{morley2019google}, so they could help rural areas and developing countries, where physician shortage is severe~\cite{jankharia2008commentary}, at relatively low cost.
\end{itemize}

\noindent \textbf{How:} Safety and feeling safe
\vspace{-0.1cm}
\begin{itemize}
\item \textbf{Challenges}: Considering multiple metrics, such as sensitivity and specificity~\cite{rossini2016diagnostic}, and dataset/task dependency~\cite{huang2018medical}, accuracy could be unreliable, so ensuring safety is challenging. Moreover, reassuring physicians and patients is important to actually use AI in a clinical environment~\cite{krittanawong2018rise}.\\
\vspace{-0.2cm}
\item \textbf{Solutions}: We should integrate various clinical data, such as blood test biomarkers and multiomics, with images~\cite{li2019medical}. Moreover, developing bias-robust technology is important since confounding factors are inevitable~\cite{li2018fully}. To prevent oversight, prioritizing sensitivity over specificity is essential while maintaining a balance~\cite{jain2017addressing}. We should also devise education for medical AI users, such as result interpretation, to reassure patients~\cite{wartman2019reimagining}.
\end{itemize}
\vspace{-0.4cm}
\subsection{Questionnaire Survey Results}
\label{sec:4.2}
\begin{figure}[t!]
  \centering
  \centerline{\includegraphics[width=0.9\linewidth]{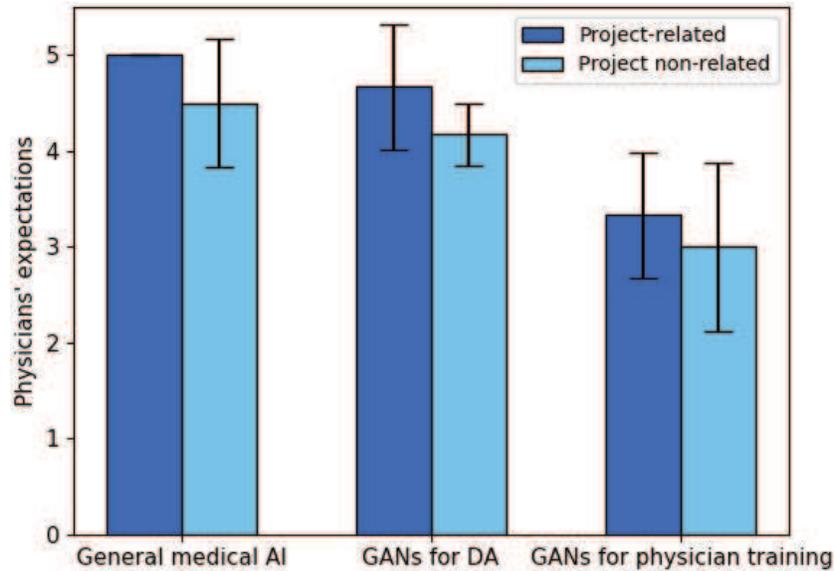}}
\caption{Bar chart of the expectations on medical AI, expressed by five-point Likert scale scores, from 9 Japanese physicians: 3 project-related physicians and 6 project non-related radiologists, respectively. The vertical rectangles and error bars denote the average scores with 95\% confidence intervals.}
\label{fig2}
\vspace{-0.4cm}
\end{figure}

We show the questions and Japanese physicians' response summaries. Concerning the following \textbf{Questions 1,2,3}, Fig.~\ref{fig2} visually summarizes the expectation scores on medical AI (i.e., general medical AI, GANs for DA, and GANs for physician training) from both 3 project-related physicians and 6 project non-related radiologists.\\

\noindent \textbf{Question 1:} Are you keen to exploit medical AI in general when it achieves accurate and reliable performance in the near future?
\vspace{-0.1cm}
\begin{itemize}
\item \textbf{Response summary}: As expected, the project-related physicians are AI-enthusiastic while the project non-related radiologists are also generally very positive about the medical AI. Many of them appeal the necessity of AI-based diagnosis for more reliable diagnosis because of the lack of physicians. Meanwhile, other physicians worry about its cost and reliability. We may be able to persuade them by showing expected profitability (e.g., currently CT scanners have an earning rate 16\% and CT scans require 2-20 minutes for interpretation in Japan). Similarly, we can explain how experts annotate medical images and AI diagnoses disease based on them (e.g., multiple physicians, not a single one, can annotate the images \textit{via} discussion).
\end{itemize}

\newpage

\noindent \textbf{Question 2:} What do you think about using GAN-generated images for DA?\vspace{-0.1cm}
\begin{itemize}


\item \textbf{Response summary}: As expected, the project-related physicians are very positive about the GAN-based DA while the project non-related radiologists are also positive. Many of them are satisfied with its achieved accuracy/sensitivity improvement when available annotated images is limited. However, similarly to their opinions on general Medical Image Analysis, some physicians question its reliability.
\end{itemize}

\noindent \textbf{Question 3:} What do you think about using GAN-generated images for physician training?\vspace{-0.1cm}
\begin{itemize}

\item \textbf{Response summary}: We generally receive neutral feedback because we do not provide a concrete physician training tool, but instead general pathology-aware generation ideas with example synthesized images---thus, some physicians are positive, and some are not. A physician provides a key idea about a pathology-coverage rate for medical student/expert physician training, respectively. For extensive physician training by GAN-generated atypical images, along with pathology-aware GAN-based extrapolation, further GAN-based extrapolation would be valuable.
\end{itemize}

\newpage

\noindent \textbf{Question 4:} Any comments/suggestions about our projects towards developing clinically relevant AI-powered systems based on your experience?\vspace{-0.1cm}
\begin{itemize}

\item \textbf{Response summary}: Most physicians look excited about our pathology-aware GAN-based image augmentation projects and express their clinically relevant requests. The next steps lie in performing further GAN-based extrapolation, developing reliable and clinician-friendly systems with new practice guidelines, and overcoming legal/financial constraints.
\end{itemize}

\vspace{-0.45cm}

\section{Conclusion}
\vspace{-0.1cm}
Our first clinically valuable AI-envisioning workshop between people with various AI and/or Healthcare background reveals the intrinsic gap between both sides and its preliminary solutions. Regarding clinical significance/interpretation, medical AI could play a key role in supporting physicians with diagnosis, therapy, and surgery. For data acquisition, countries should utilize their unique medical environment, such as complete medical checkup for Japan. Commercial deployment could come as AI-powered hospitals and medical manufacturers' AI service. To assure safety and feeling safe, we should integrate various clinical data and devise education for medical AI users. We believe that such solutions on \textit{Why} and \textit{How} would play a crucial role in connecting inter-disciplinary research and clinical applications. 

Through a questionnaire survey for physicians, we confirm our pathology-aware GANs' clinical relevance for diagnosis as a clinical decision support system and non-expert physician training tool: many physicians admit the urgent necessity of general AI-based diagnosis while welcoming our GAN-based DA to handle the lack of medical images. Thus, GAN-powered physician training is promising only under careful tool designing.

We find that better DA and expert physician training require further generation of images with abnormalities. Therefore, for better GAN-based extrapolation, we plan to conduct (\textit{i}) generation by parts with coordinate conditions~\cite{lin2019coco}, (\textit{ii}) generation with both image and radiogenomic conditions~\cite{xu2019correlation}, and (\textit{iii}) transfer learning among different body parts and disease types. Whereas this paper only explores the Japanese medical context and pathology-aware GANs, our findings are more generally applicable and can provide insights into the clinical practice in other countries.

\vspace{-0.3cm}

%
%
%
\bibliographystyle{splncs}
 \bibliography{samplepaper}
\end{document}